\documentclass[final]{cvpr}

\usepackage{times}
\usepackage{epsfig}
\usepackage{graphicx}
\usepackage{amsmath}
\usepackage{amssymb}

\usepackage[pagebackref=true,breaklinks=true,colorlinks,bookmarks=false]{hyperref}

\def\cvprPaperID{63} 
\def\confYear{CVPR 2021}

\begin{document}

\title{Multi-scale Self-calibrated Network for Image Light Source Transfer}

\author{Yuanzhi Wang, Tao Lu \thanks{indicates corresponding author.}, Yanduo Zhang, Yuntao Wu\\
Hubei Key Laboratory of Intelligent Robot, Wuhan Institute of Technology, Wuhan, China, 430073.\\
{\tt\small \{w906522992, lutxyl\}@gmail.com, zhangyanduo@hotmail.com, ytwu@wit.edu.cn}
}

\maketitle

\begin{abstract}
   Image light source transfer (LLST), as the most challenging task in the domain of image relighting, has attracted extensive attention in recent years.
   In the latest research, LLST is decomposed three sub-tasks: scene reconversion, shadow estimation, and image re-rendering, which provides a new paradigm for image relighting. 
   However, many problems for scene reconversion and shadow estimation tasks, including uncalibrated feature information and poor semantic information, are still unresolved, thereby resulting in insufficient feature representation.
   In this paper, we propose novel down-sampling feature self-calibrated block (DFSB) and up-sampling feature self-calibrated block (UFSB) as the basic blocks of feature encoder and decoder to calibrate feature representation iteratively because the LLST is similar to the recalibration of image light source.
   In addition, we fuse the multi-scale features of the decoder in scene reconversion task to further explore and exploit more semantic information, thereby providing more accurate primary scene structure for image re-rendering.
   Experimental results in the VIDIT dataset show that the proposed approach significantly improves the performance for LLST.
   Codes have been released at \url{https://github.com/mdswyz/MCN-light-source-transfer}.
\end{abstract}

\thispagestyle{empty} 

\section{Introduction}
Transferring the current light source setting of given image to the target light source setting is a domain-specific image relighting task, which have many potential applications in data augmentation, photo editing, and gaming industry.
Inappropriate light source usually causes various visual degradation problems, such as undesired shadows, distorted colors, and unrealistic textures.

Some early methods have been proposed that to reduce the degradation caused by improper illumination conditions. 
For example, Wu \emph{et al.} \cite{HDR} improved the image quality by increasing the dynamic range of the low-contrast regions, which can be regarded as a refinement of local light conditions. 
Retinex-based methods \cite{RetinexNet} assumed that the images can be decomposed reflectance and illumination components, where
the reflectance component stored the inherent scene structure that is unchangeable in different illumination conditions.
By refining the global illumination, it can improve the visual quality of the images.

Despite the great success in image relighting, aforementioned methods can not manipulate the light source settings (e.g., direction, color temperature) to achieve more realistic image quality. 
Compared with the traditional image relighting task, image light source transfer (LLST) is an extremely challenging task because it is difficult to remove and recast shadows, and it is also difficult to relight and re-darken specific regions, which increases the difficulty of producing satisfactory image.

In view of the above problems, some researchers have explored the manipulation and transfer of image light source in latest VIDIT dataset \cite{vidit}.
Puthussery \emph{et al.} \cite{WDRN} proposed a encoder-decoder network with discrete wavelet transform based decomposition and transferred an image from a source illumination setting to a target setting.
The YorkU team \cite{AIM2020} defined the task as two parts: image normalization and image relighting, and used two encoder-decoder networks to infer the target results.
Wang \emph{et al.} \cite{DRN} designed a deep relighting network with three parts: scene reconversion, shadow prior estimation, and re-renderer.
Scene reconversion part aimed to restore the primary scene structure, shadow prior estimation part predicted light effect from the target light direction, and re-renderer part combined the primary structure with the recast shadow to reconstruct target light source setting.
This novel paradigm had achieved the best PSNR in the ``AIM2020 - One-to-one relighting challenge'' \cite{AIM2020}.

However, since LLST is a task of recalibrating the light source settings \cite{WDRN}, the above method ignores the effective calibration of the feature information during feature extraction, resulting in insufficient feature representation, and rendering unsatisfactory results ultimately.

\begin{figure*}[t]
	\centering{\includegraphics[width=\linewidth]{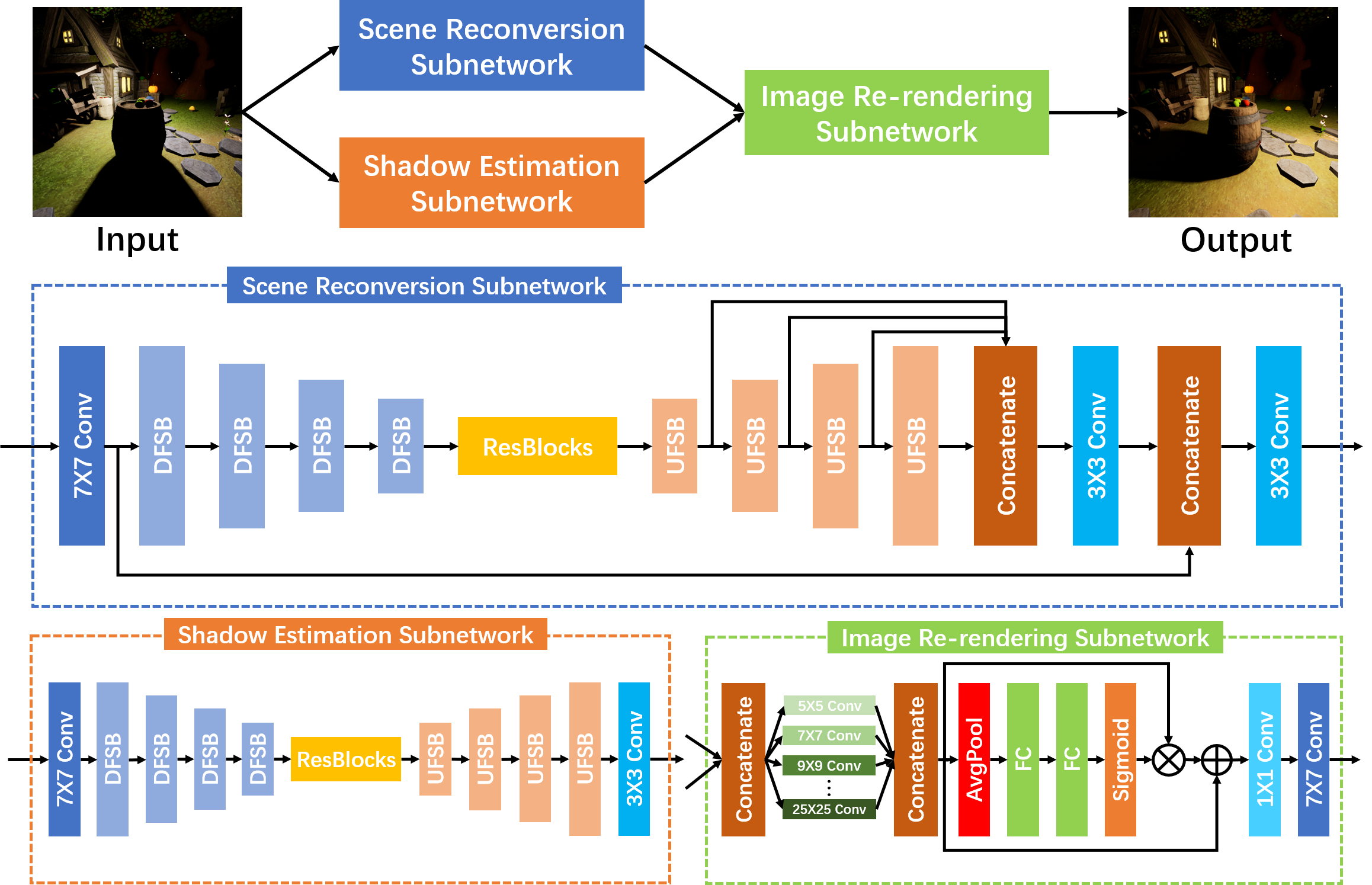}}
	\caption{Overall architecture of the proposed method includes three parts: scene reconversion subnetwork, shadow estimation subnetwork, and image re-rendering subnetwork.}
	\label{fig:network}
\end{figure*}

To address the above issues effectively, we design novel down-sampling feature self-calibrated block (DFSB) and up-sampling feature self-calibrated block (UFSB) as the basic blocks of feature encoder and decoder in scene reconversion and shadow estimation tasks.
By generating the input feature weight of each basic block, the output feature of each block is calibrated iteratively.

In addition, in order to further explore and exploit more semantic information, we fuse the multi-scale features of the decoder in scene reconversion task to better reveal primary scene structure of input image, thereby re-rendering better results with recast shadow.
The proposed method focuses on transferring the direction and color temperature of the light source. Experimental results show that our method exceed the current state-of-the-arts.
The contributions of the proposed method can be summarized as follows:

(\romannumeral1) Considering that LLST is a task of recalibrating the light source settings, we propose novel DFSB and UFSB as the basic blocks of feature encoder and decoder to calibrate feature information for scene reconversion and shadow estimation tasks iteratively, thereby improving feature representation ability.

(\romannumeral2) To further explore and exploit more semantic information, we design the multi-scale feature fusion method to the feature decoder structure of scene reconversion task, which provides more accurate primary scene structure for image re-rendering task.

\section{Related Work}
Image relighting is a popular topic in computer vision. 
In some early studies, researchers have often worked on enhancing low-light images.
Histogram equalization \cite{adaptiveHE} restrain the histograms of the output images to increases the discrimination of the low-contrast regions.
The high-dynamic-range methods \cite{HDROLD} can increase the dynamic range of the low-contrast regions and refine local contrast, thereby improving the image quality.
Retinex theory \cite{retinex} decomposes the input image into the reflection component and the illumination component, and the illumination of the image can be changed by adjusting the illumination component.

Convolutional neural networks (CNNs) has achieved great success in computer vision recently due to its powerful feature representation capacity, which has been widely used in a variety of upstream and downstream tasks, like image super-resolution \cite{GLFSR,DPDFN}, image defogging \cite{Defogging}, object detection \cite{FasterRCNN,SPMF}, etc.
Considering the advantages of CNNs, the CNNs-based image relighting methods have also made considerable progress.
Lore \emph{et al.} \cite{LLNet} designed an auto-encoder architecture for low-light image enhancement.
Wei \emph{et al.} \cite{RetinexNet} combines CNNs with retinex theory to improve the performance of image relighting significantly.
Jiang \emph{et al.} \cite{Enlightengan} proposed an unsupervised generative adversarial network (GAN) to relight the whole image.

Although the above studies have achieved satisfactory performance in low light image enhancement, these methods can only adjust the brightness of the image globally or locally.
In recent studies, some researchers manipulate and transfer light source on portrait scenes \cite{portraitrelighting2,facerelighting}.
These methods require some prior information (e.g., face landmarks, geometric priors) that can not be obtained in general scenes.
In the latest research, researchers began to explore the light source transfer in general scenes due to the release of VIDIT dataset \cite{vidit} that is a novel virtual image dataset for light source transfer in general scenes.
Das \cite{MSRNet} proposed a multi-scale relighting network to transfer an image from a original light source setting to a target light source setting.
The YorkU team \cite{AIM2020} defined the task as two parts: image normalization and image relighting, and used two encoder-decoder networks to infer the target results.
Wang \emph{et al.} \cite{DRN} decomposed LLST into three parts: scene reconversion, shadow prior estimation, and re-renderer.
This novel paradigm achieved satisfactory performance.

However, a vital issue was ignored by previous work. Since LLST is similar to the image light source recalibration task, it is necessary to calibrate the features effectively during the feature encoding and decoding to obtain the better feature representation. 
The proposed method creatively designs DFSB and UFSB to calibrate the feature representation effectively.

\section{Image light source transfer via multi-scale self-calibrated network}

Similar to \cite{DRN}, the proposed method consists of three parts: scene reconversion subnetwork, shadow estimation subnetwork, and image re-rendering subnetwork, the overall  architecture of the proposed method is shown in Figure \ref{fig:network}.
Firstly, the input image is processed in the scene reconversion subnetwork to extract primary scene structures by removing the light effects.
At the same time, the shadow estimation subnetwork 
aims to the change of the lighting effect, which recasts shadows according to the target light source setting.
Finally, the image re-rendering subnetwork learns the target color temperature and perceives the global light source effect, and re-renders the image with the support of the primary scene structure information and the predicted shadow.

Both the scene reconversion subnetwork and shadow estimation subnetwork have a similar deep auto-encoder structure and use the proposed DFSB and UFSB as the basic blocks for feature encoding and decoding. In addition, we fuse multi-scale features in the feature decoding part of the scene reconversion subnetwork to enrich the semantic information of the output features. 
For the image re-rendering subnetwork, we use the re-renderer component proposed in the previous work \cite{DRN} as the backbone.

\subsection{Modeling of image light source transfer}
LLST is a challenging low-level vision task that focuses on reconstructing input image $X$ (under any light
source $\sigma$) with the target light source setting $\theta$. Inspired by the previous work \cite{retinextheory,RetinexNet,DRN}, we model that the images can be decomposed into two components, primary scene structure $P$ that is unchangeable in different light conditions and light source setting $\sigma$ that provides global illumination , shadow effects, and color temperature.
The input image can be formulated as:
\begin{equation}
X = H_{\sigma}(P),
\end{equation}
where $H_{\sigma}(.)$ denotes a lighting operation.
To re-render the image $X$ with light source setting $\theta$, it firstly needs to remove $\sigma$, reconverting the primary scene structure $P$ from the input image $X$.
Then, with the target lighting operation $H_{\theta}(.)$, the image $Y$ with the target light source setting can be obtained by:
\begin{equation}
Y = H_{\theta}(P) = H_{\theta}(H_{\sigma}^{-1}(X)).
\end{equation}
The key point of the reconversion process $H_{\sigma}^{-1}(.)$ is to eliminate the shadows, while the lighting operation $H_{\theta}(.)$ is to recast shadows for the target light source setting.
However, it is difficult to construct the lighting operation $H_{\theta}(.)$ because the geometric information is unavailable in the single image relighting task.
Hence, instead of finding the lighting operation $H_{\theta}(.)$ directly, the proposed method aims to construct a transferring operation $H_{(\sigma\to\theta)}(.)$ that transfer the target light effects (e.g., shadows, light source direction, color temperature) from the input image to target image. 
The above transferring process can be formulated as:
\begin{equation}
Y = H_{re-rendering}(H_{\sigma}^{-1}(X),H_{(\sigma\to\theta)}(X)),
\end{equation}
where the $H_{re-rendering}(.)$ denotes a a re-rendering process.

\begin{figure}[t]
	\centering{\includegraphics[width=\linewidth]{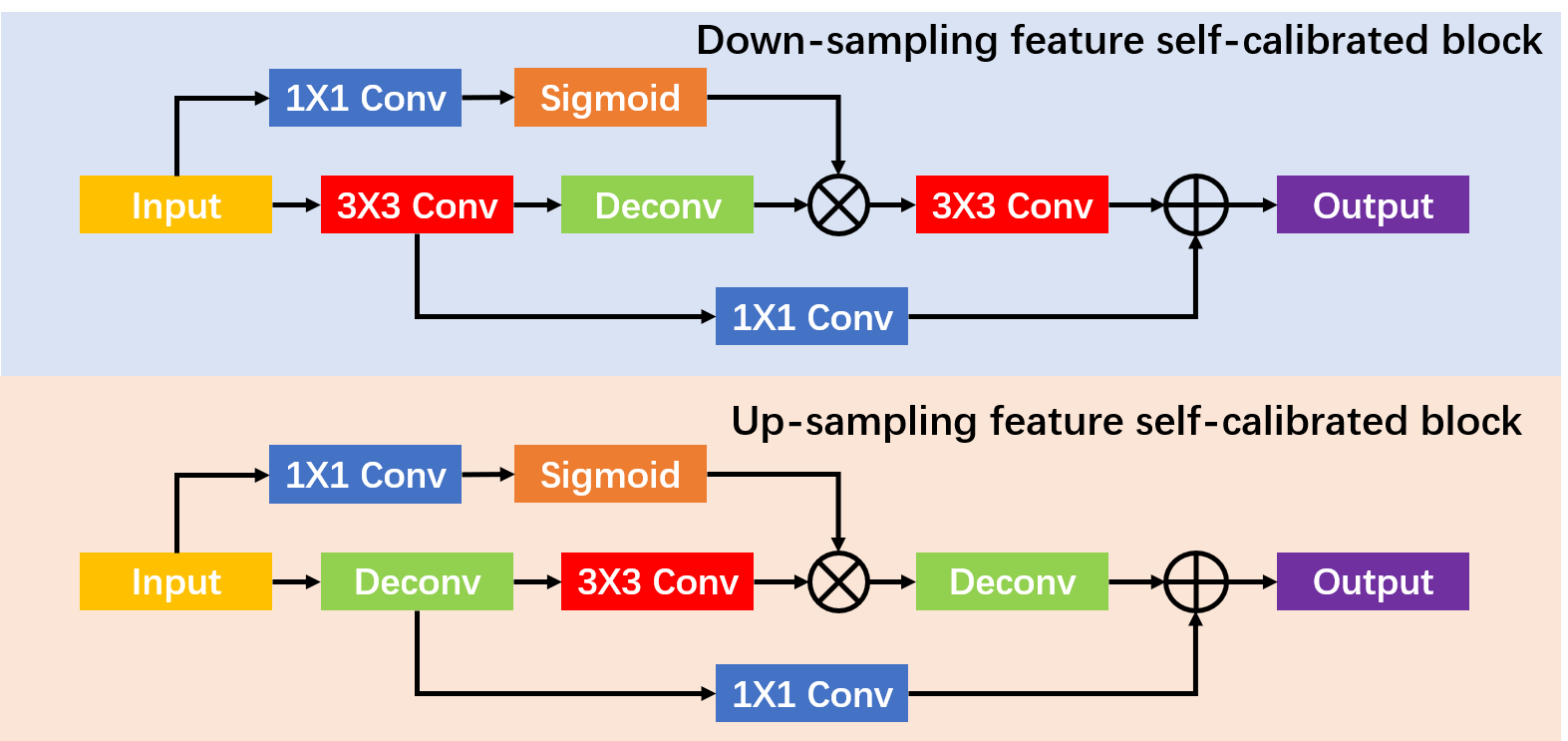}}
	\caption{Architecture of DFSB (top) and UFSB (bottom)}
	\label{fig:SC}
\end{figure}
\subsection{Feature self-calibrated block for feature encoding and decoding}
The objective of the LLST is to transfer the light effects from the input image to target image, which is similar to the recalibration of image light source.
Hence, we propose novel DFSB and UFSB as the basic blocks of feature encoder and decoder in scene reconversion and shadow estimation subnetworks to calibrate feature representation iteratively.
Figure \ref{fig:SC} shows the architecture of proposed DFSB and UFSB, which consist of encoding and decoding operations that map the feature information between the input and latent spaces.

To take the DFSB for example, it firstly maps the input feature $F_{input}$ to small-scale space through a 3$\times$3 convolution layer with stride of 2. The small-scale feature $F_{lr}$ can be formulated as:
\begin{equation}
F_{lr} = H_{3\times3 Conv}(F_{input}),
\end{equation}
where $H_{3\times3 Conv}(.)$ denotes a 3$\times$3 convolution layer.
Then, a 4$\times$4 deconvolution layer with stride of 2 maps $F_{lr}$ back to the input scale space, which is formulated as:
\begin{equation}
F_{hr} = H_{4\times4 Deconv}(F_{lr}),
\end{equation}
where $H_{4\times4 Deconv}(.)$ denotes a 4$\times$4 deconvolution layer.
At the same time, another branch generates calibration weight $weight_{calibration}$ through a 1$\times$1 convolution layer and $Sigmiod$ function. $weight_{calibration}$ is formulated as:
\begin{equation}
weight_{calibration} = Sigmiod(H_{1\times1 Conv}(F_{input})),
\end{equation}
where $H_{1\times1 Conv}(.)$ denotes a 1$\times$1 convolution layer.
This $weight_{calibration}$ is multiplied by $F_{hr}$ to obtain the calibrated feature $F_{hr}^{calibration}$ with input scale space.
$F_{hr}^{calibration}$ is formulated as:
\begin{equation}
F_{hr}^{calibration} = weight_{calibration} \times F_{hr}.
\end{equation}
Finally, the feature $F_{hr}^{calibration}$ is remapped to a small-scale space by a 3$\times$3 convolution layer and fused with $F_{lr}$ through an element-wise summation operation to obtain the output feature $F_{output}$. $F_{output}$ is formulated as:
\begin{equation}
F_{output} = H_{Sum}[H_{3\times3 Conv}(F_{hr}^{calibration}), F_{lr}],
\end{equation}
where $H_{Sum}[.]$ denotes the element-wise summation operation.

Similar to the above process, the whole process of UFSB can be formulated as:
\begin{equation}
F_{output} = H_{Sum}[H_{4\times4 Deconv}(F_{lr}^{calibration}), F_{hr}].
\end{equation}

\subsection{Scene reconversion subnetwork}
The purpose of the scene reconversion subnetwork is to extract the primary scene structure information from the input image so that the light effects should be removed.
As shown in Figure \ref{fig:network}, the proposed subnetwork refers to the auto-encoder structure \cite{autoencoder}.
The feature encoding and decoding parts of the whole subnetwork use DFSB and UFSB as basic blocks, and fuse the multi-scale features in decoding parts to enrich semantic information.

Firstly, we use one convolutional layer with a kernel size of 7$\times$7 to extract the shallow feature $F_{shallow}$ from the input images.
Then, the $F_{shallow}$ is downsampled by the DFSB four times to find the discriminative features for the scene.
The channels are doubled after each downsampling process to preserve feature information as much as possible. The encoded feature is formulated as:
\begin{equation}
F_{encoder} = H_{DB}^{4}(H_{DB}^{3}(H_{DB}^{2}(H_{DB}^{1}(F_{shallow})))),
\end{equation}
where $H_{DB}(.)$ and $F_{encoder}$ denote the function of DFSB and output of fourth DFSB, respectively.
After feature encoding, we design a ResBlocks module, which consists of nine residual blocks, to remove the light effects.
The removed feature $F_{Res}$ is formulated as:
\begin{equation}
F_{Res} = H_{Res}(F_{encoder}),
\end{equation}
where $H_{Res}(.)$ denotes the ResBlocks module.
Next, four UFSBs upsample the feature map back to the original size gradually. 
\begin{equation}
F_{decoder} =H_{UB}^{4}(H_{UB}^{3}(H_{UB}^{2}(H_{UB}^{1}(F_{Res})))),
\end{equation}
where $H_{UB}(.)$ and $F_{decoder}$ denotes the function of UFSB and the output of fourth UFSB, respectively; and the outputs of the other three UFSBs are denoted as $F_{UFSB}^{1}$, $F_{UFSB}^{2}$, and $F_{UFSB}^{3}$, respectively.
Meanwhile, we fuse multi-scale features from first to fourth UFSB into a single one. 
Specifically, the deconvolution layers are adopted to make multi-scale features with the same resolution as $F_{decoder}$, and these multi-scale features are concatenated along the channel dimension.
The concatenated feature $F_{concat}$ is formulated as:
\begin{equation}
F_{up}^{1} = H_{8\times8 Deconv}(F_{UFSB}^{1}),
\end{equation}
\begin{equation}
F_{up}^{2} = H_{4\times4 Deconv}(F_{UFSB}^{2}),
\end{equation}
\begin{equation}
F_{up}^{3} = H_{4\times4 Deconv}(F_{UFSB}^{3}),
\end{equation}
\begin{equation}
F_{concat} = H_{c}[F_{up}^{1}, F_{up}^{2}, F_{up}^{3}, F_{decoder}],
\end{equation}
where $H_{c}[.]$ denotes the concatenating feature operation along the channel dimension.
$F_{concat}$ contains rich semantic information, which lays a solid foundation for image re-rendering.
Finally, in order to obtain rich semantic information while maintaining structural information, we concatenate $F_{concat}$ with shallow feature $F_{shallow}$ by a skip connection, and reduces the channels through a convolution layer from 64 to 32 as the input of image re-rendering subnetwork.

To train the scene reconversion subnetwork, it need primary scene structure image as the ground-truth.
However, the primary scene structure image is difficult to define because we have only the target image. 
Inspired by the previous work, we use the shadow-free image generated by exposure fusion method \cite{exposure} as the ground-truth, which is provided by \cite{DRN}.
For the scene reconversion subnetwork, the shadow-free image is reconstructed by a convolutional layer, which transfers the latent feature space back to the image space. 
In the training phase, a discriminator is attached to assist the training of the scene reconversion subnetwork. The discriminator structure is proposed in \cite{GAN} to extract the global representations hierarchically.
The detailed parameter settings of the discriminator and loss function are
reported in \cite{DRN}.

\subsection{Shadow estimation subnetwork}
To produce the light effects from the target light source, we design a shadow estimation subnetwork with the architecture as shown in Figure \ref{fig:network}, which is similar to the proposed scene reconversion subnetwork.
Compared with the scene reconversion network, the shadow estimation subnetwork discards the skip connection and multi-scale feature fusion, which makes the network pay more attention to the global light effect. 

To train the shadow estimation subnetwork, the ground-truth is the image under the target light source setting.
In the training phase of network, in order to make the network focus on shadow regions, we add an additional shadow discriminator, which uses the same structure as the previous discriminator.
Specifically, it is firstly rectified to give focus to the low-intensity regions (such as dark regions, shadow regions) by $z = \min(x,y)$, where the $y$ denotes the estimated pixel intensity. The $z$ represents rectified value that is inputted to the discriminator.
The $x$ is a hyperparameter to pre-define threshold for the sensitivity of the shadows. 
Referring to the previous work \cite{DRN}, it is set to $0.059 = 15 / 255$.

\subsection{Image re-rendering subnetwork}
After the processing of the scene reconversion and shadow estimation tasks, the predicted primary scene structure and light effects are fused to re-render the output image with target light source setting.
Figure \ref{fig:network} shows structure of the image re-rendering subnetwork, we firstly uses several convolution layer with different kernel sizes from 3$\times$3 to 25$\times$25 to utilize the information of different perception scales, which extracts rich multi-scale features for the subsequent process.
Second, these features are concatenated into a single one. 
Third, the concatenated feature is recalibrated by a recalibration module, which is similar to \cite{SENet}, to explore the recalibration weights for different scale space.
Finally, a convolutional layer with kernel size of 7$\times$7 re-render the recalibrated feature from the feature space to the image space.

\begin{figure*}[t]
	\centering{\includegraphics[width=\linewidth]{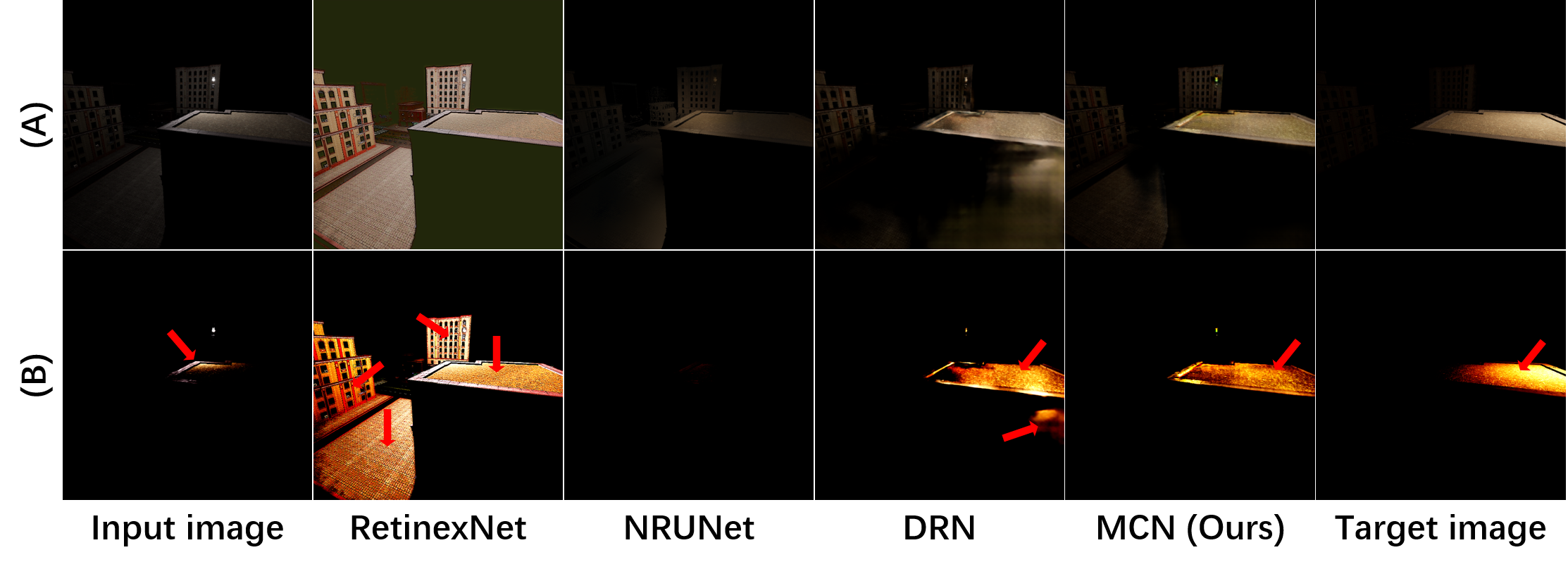}}
	\caption{Visual results on a selected test sample (image with ID ``300'' of the validation dataset). The images in row (A) include the input image, target image, and the results inferred by different methods. The images in row (B) are the results after contrast enhancement to highlight the light effects and the red arrows indicate the directions of the light source.}
	\label{fig:simple}
\end{figure*}

\section{Experiments}

\subsection{Dataset and Implementation Details}
The performance of our method is evaluated on the novel VIDIT (Virtual Image Dataset for Illumination Transfer) dataset \cite{vidit}.
The VIDIT dataset contains 390 different virtual scenes, where there are 300 scenes for training, 45 scenes for validation and 45 scenes for testing separately.
Each scene is captured with 40 different light source settings, which are all the combinations of 5 color temperatures (2500K, 3500K, 4500K, 5500K, and 6500K) and 8 light source directions (N, NE, E, SE, S, SW, W, NW).
All scenes are rendered with a resolution of 1024$\times$1024 pixels, include metal, wood, water, etc.
We participated the ``NTIRE 2021 Depth Guided Relighting Challenge Track 1: One-to-one relighting'' \cite{2021ntire}. 
The objective of competition is that, given an input image under any light source settings, the method should give the result under a specific light source setting (color temperature is 4500k and light source direction is from east).
We use all possible pairs from the 300 training scenes to train the network, and the provided 45 validation images for evaluation.

Our model is trained by Adam optimizer with the momentum of 0.5 and learning rate of $2\times10^{-4}$, and the training images were resized from 1024$\times$1024 to 512$\times$512.
All experiments were conducted through PyTorch with two NVIDIA RTX 2080Ti GPUs. 

\subsection{Evaluation Metrics}
The results are evaluated with standard evaluation metrics like Peak Signal to Noise Ratio (PSNR) and Structural Similarity (SSIM) \cite{SSIM}.
Furthermore, we use novel perceptual metrics like Learned Perceptual Image Patch Similarity (LPIPS) \cite{LPIPS} and Mean Perceptual Score (MPS) \cite{AIM2020}.
LPIPS measures the perceptive quality of image, in which a smaller value means more perceptual similarity.
MPS is the average of the SSIM and LPIPS, which is formulated as:
\begin{equation}
MPS = 0.5\times(SSIM+(1-LPIPS)).
\end{equation}
The above four evaluation metrics ensure the credibility of the evaluation results.

\subsection{Effect of the proposed self-calibrated block and multi-scale feature fusion}
In this subsection, we conduct the ablation studies on the VIDIT dataset to demonstrate the effectiveness of self-calibrated block and multi-scale feature fusion.

First, we remove all feature self-calibration processes of UFSBs and DFSBs, and remove the multi-scale feature fusion on the scene reconversion subnetwork to form the first experiment.
Then we add the feature self-calibration process of UFSBs and DFSBs to the first experiment to form the second experiment.
Finally, we add multi-scale feature fusion in the scene reconversion subnetwork to form the third experiment based on the first experiment.
Table \ref{tab:1} shows the quantitative results of the above experiments, we can conclude that the feature self-calibration and multi-scale feature fusion can lead to considerable performance improvements, and the best performance is achieved when both are used.

\begin{table}[h]
	\begin{center}
		\caption{Quantitative comparison of different methods. MCN as an abbreviation for our proposed method. Cal and MS denote feature self-calibration and multi-scale feature fusion, respectively. The best results are \textbf{highlighted}. }\label{tab:1}
		\begin{tabular}{c|c|c|c}
			\hline
		    Methods & PSNR & SSIM & LPIPS\\
		    \hline
		    MCN without Cal and MS & 17.55& 0.6206& 0.4087\\
		    MCN without Cal & 17.64& 0.6234& 0.4091\\
		    MCN without MS & 17.80 & 0.6407& 0.3827\\
		    MCN & \textbf{17.86} & \textbf{0.6487} & \textbf{0.3794}\\
			\hline 
		\end{tabular}
	\end{center}
\end{table}

\begin{figure*}[t]
	\centering{\includegraphics[width=14cm]{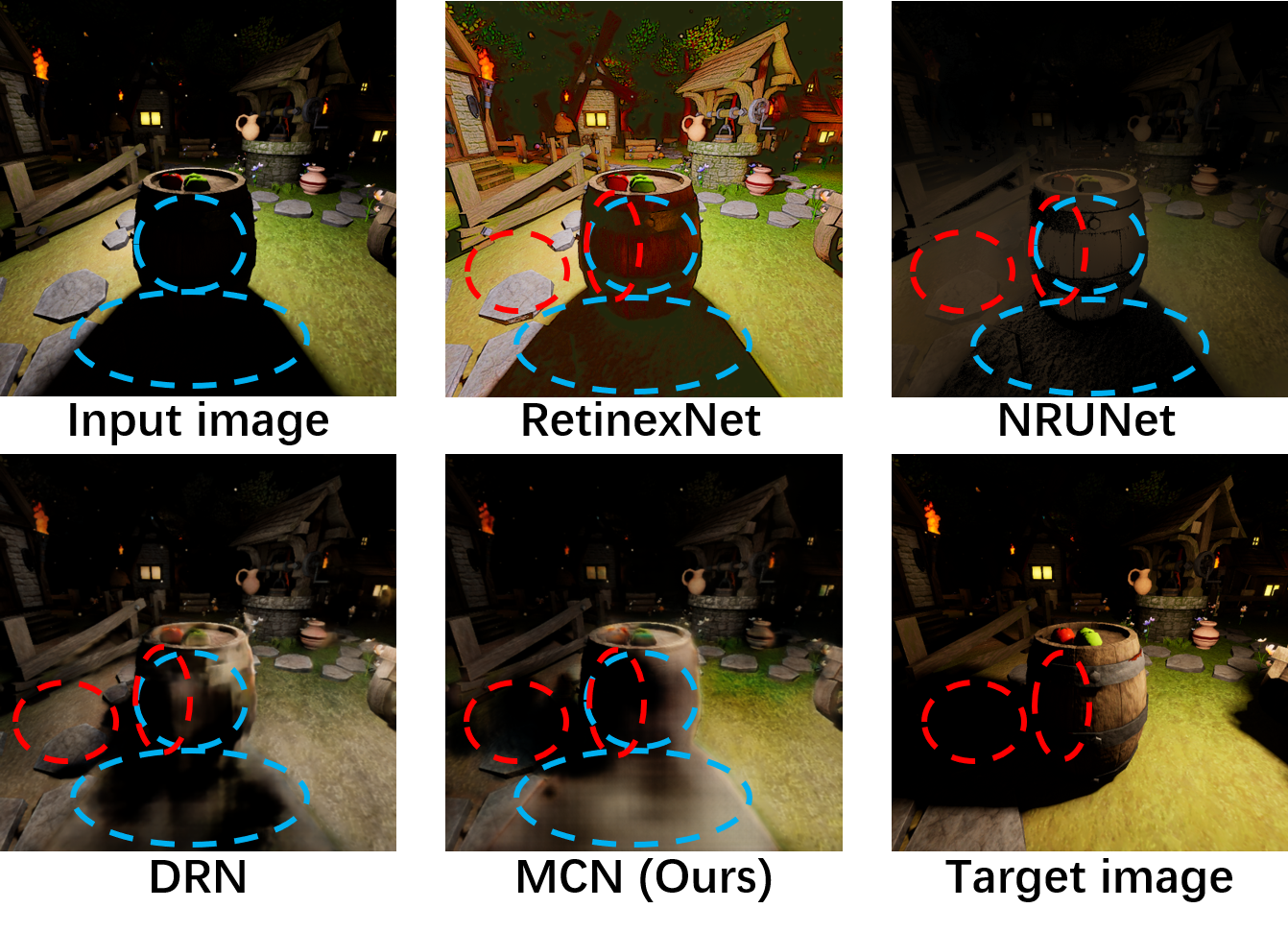}}
	\caption{Visual results on a selected challenging  test sample (image with ID ``318'' of the validation dataset). The blue circles represent the shadow regions of the input image, and the red circles represent the shadow regions of the target image. Compared to other methods, our method can effectively remove the shadows of the input image and recast the shadows of the target image.}
	\label{fig:hard}
\end{figure*}

\subsection{Comparison with State-of-the-Arts}
Since LLST is a novel topic in the domain of image relighting, there are only two methods that can be publicly used for our comparison currently, namely DRN \cite{DRN} and NRUNet \cite{AIM2020}.
Besides comparing with the above two methods, we have also made comparison with RetinexNet \cite{RetinexNet}, which is a representative relighting method based on Retinex theory.
We retrained these above methods in the unified dataset with publicly available codes provided by authors, and the comparison was made using the VIDIT validation dataset.

Table \ref{tab:2} lists the quantitative results of different methods, it is obviously that our method outperforms state-of-the-arts in PSNR, SSIM, LPIPS, and MPS significantly.
These quantitative results demonstrate that the proposed method can better re-render the target light source setting and obtain better perceptual similarity through calibrating features iteratively and fusing multi-scale features.

\begin{table}[h]
	\begin{center}
		\caption{Comparison with the state-of-the-arts on VIDIT validation dataset. The best results are \textbf{highlighted}. }\label{tab:2}
		\begin{tabular}{c|c|c|c|c}
			\hline
			Methods & PSNR & SSIM & LPIPS& MPS\\
			\hline
			RetinexNet \cite{RetinexNet} & 12.15 & 0.1659 & 0.5686& 0.2987 
			\\
			NRUNet \cite{AIM2020} & 16.24 & 0.5805 & 0.4147 & 0.5829 
			\\
			DRN \cite{DRN} & 17.59 & 0.6088 & 0.3920 & 0.6084 
			\\
			MCN (Ours) & \textbf{17.86} & \textbf{0.6487} & \textbf{0.3794}& \textbf{0.6347 
			}\\
			\hline 
		\end{tabular}
	\end{center}
\end{table}

In terms of visual performance, Figure \ref{fig:simple} shows the qualitative results of these methods.
Row (A) in Figure \ref{fig:simple} shows the input image, target image, and the re-rendering results of each method.
In order to highlight the lighting effects, we perform contrast enhancement on all images in row (A) to generate images in row (B), and use red arrows to mark all the light source directions in each image.
From the Figure \ref{fig:simple} we can find that RetinexNet can not change the color temperature of the image and the direction of the light source because it only considers enhancing the global illumination.
NRUNet can change to correct color temperature for the target light source, but it fails to transfer the light direction, and the light source disappears in the contrast-enhanced image.
The possible reason for the disappearance of the light source is that NRUNet fails to re-render the target light source settings well after normalizing the image.
The DRN can transfer to correct light source setting but produces some unrealistic artifacts, which lead to additional light source in the image and degrade the visual quality.
In contrast, our method re-renders correct light direction and color temperature with a good perceptual quality.

The scene of the above test sample is relatively simple. To further demonstrate the effectiveness of the proposed method, we select a challenging test sample in the validation dataset for experiment.
Figure \ref{fig:hard} shows the visual results on the challenging test sample.
From the input image and the target image, it can be concluded that the difficulty of this scene is to remove the shadow regions of the input image and recast the shadow regions of the target image.
To clearly distinguish the differences between the visual results of different methods, we first mark the shadow regions on the input image (blue circles) and the target image (red circles), respectively.
Then, the results obtained by different methods are marked with the shadow regions of the input image (blue circles) and the shadow regions of the target image (red circles).
The purpose of this is to show that the shadow in the blue circle region need to be removed, while the shadow in the red circle region need to be recast.

From the Figure \ref{fig:hard} we can conclude that RetinexNet, NRUNet, and DRN can not remove the shadows of the input image and recast the shadows for the target regions.
In contrast, the proposed method not only removes the shadows of the input image, but also recasts the shadows of the target image, which benefits from the effective calibration of feature representation by the proposed DFSB and UFSB.

\section{Conclusion}
In this paper, a novel multi-scale self-calibrated network is proposed to solve the problem of uncalibrated features and poor semantic information in LLST task.
We design DFSB and UFSB to calibrate the features effectively, and design multi-scale feature fusion to explore and exploit more semantic information, thereby improving the LLST performance from the input light source setting to the target light source setting.
However, our proposed method also has some limitations.
For example, after removing the shadow regions of the input image, the texture information of the these regions can not be rendered well, resulting in the visual quality degradation.
Therefore, this point will be the focus of recent research.

\section{Acknowledgments}
This work has been supported by the National Natural Science Foundation of China (62072350, 61771353), Hubei Technology Innovation Project (2019AAA045), the Central Government Guides Local Science and Technology Development Special Projects (2018ZYYD059), 2020 Hubei Province High-value Intellectual Property Cultivation Project, the Wuhan Enterprise Technology Innovation Project (202001602011971), the Graduate Education Innovation Fund of Wuhan Institute of Technology (CX2020223).

{\small
\bibliographystyle{ieee_fullname}
\bibliography{egbib}
}

\end{document}